\documentclass{article}
\usepackage{amsmath,epsfig}
\usepackage[preprint]{spconfa4}
\usepackage{hyperref}

\let\OLDthebibliography\thebibliography
\renewcommand\thebibliography[1]{
  \OLDthebibliography{#1}
  \setlength{\parskip}{0pt}
  \setlength{\itemsep}{0pt plus 0.3ex}
}

\begin{document}\sloppy

\title{VirtualConductor: Music-driven Conducting Video Generation System}
%
\name{Delong Chen, Fan Liu*, Zewen Li, Feng Xu}
\address{College of Computer and Information, Hohai University, China\\fanliu@hhu.edu.cn}

\maketitle

\begin{abstract}
In this demo, we present the \textit{VirtualConductor}, a system that can generate conducting video from any given music and a single user's image. First, a large-scale conductor motion dataset is collected and constructed. Then, we propose Audio Motion Correspondence Network (AMCNet) and adversarial-perceptual learning to learn the cross-modal relationship and generate diverse, plausible, music-synchronized motion. Finally, we combine 3D animation rendering and a pose transfer model to synthesize conducting video from a single given user's image. Therefore, any user can become a virtual conductor through the system.
\end{abstract}

\section{Introduction}
\label{sec:intro}
    In recent years, deep learning has shown its advantages in learning discriminative feature representations \cite{ZechaoPAMI} and learning high-quality generation \cite{ACMTrans} from massive data. As a notable research line in this field, learning the cross-modal mapping from sound to human motion has drawn a lot of attention. Various types of applications, including speech gesture generation and musical gesture generation (dancing and instrument playing), have been developed in recent years. But researchers pay little attention to the motion generation of an orchestral conductor. Moreover, there is not a large-scale conductor motion dataset currently available. Therefore, we build a system to make the first attempt towards music-driven conductor motion generation and realize a virtual conductor.
    
    To build a large-scale conductor motion dataset, we first collect concert performance video recordings, then extract conductor motion by pose estimation \cite{Alphapose}. Meanwhile, different types of audio features, including MFCC, spectral centroid, spectral bandwidth, onset envelope, estimated tempo, and predominant local pulse, are extracted. Finally, the constructed dataset consists of conductor motion data and aligned music features in a total of 40 hours.
    
    However, modeling conductor motion still has several challenges. First, the conductor motion is highly complicated because it conveys various types of information, including tempo, strength, and emotion. Meanwhile, the generated motion should be closely synchronized with music. Moreover, because of different conducting styles, mapping music to conductor motion is a one-to-many task, which is difficult to learn by standard mean squared error (MSE) regression.  
    In this demo, based on the constructed dataset, we propose the \textit{VirtualConductor} system to tackle the above difficulties. We use a combination of MSE loss, pose perceptual loss, and adversarial loss to train the motion generator. In this way, the generated motion can be simultaneously diverse, plausible, and synchronized to music. Finally, by combining 3D animation rendering and pose transfer \cite{LiquidWarpingGAN} module, the system can generate conducting video from given music and a single user's image. In the following sections, we will introduce our system in detail.

\begin{figure}
    \centering
    \includegraphics[width=0.475\textwidth]{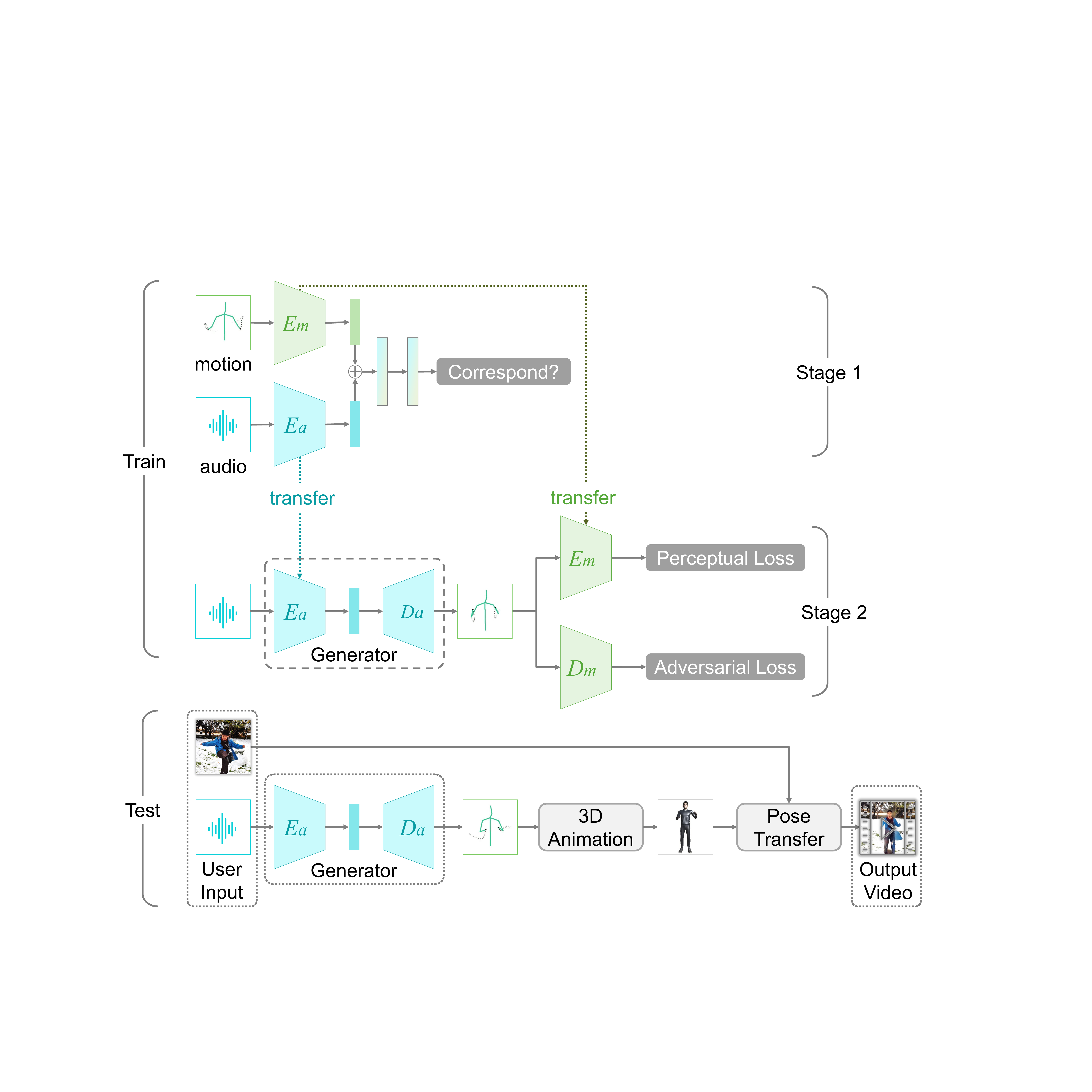}
    \caption{The pipeline of presented demo \textit{VirtualConductor}.}
    \label{fig:model_structure}
\end{figure}

\section{System Design and Implementation}
\label{sec:Approach}

\subsection{Audio Motion Correspondence Learning}
    We first design an AMCNet to learn the correspondence between audio and motion. As shown in Fig.\ref{fig:model_structure}, the AMCNet consists of a music encoder $E_a$, a motion encoder $E_m$, and fuse layers. The features extracted by two encoders are concatenated and passed to fuse layers. The AMCNet output a possibility in the range of (0,1), indicating whether the input music frame and motion frame correspond. The loss function of the AMCNet is shown in Eq.\ref{lossAMC}, where $x_i$ and $y_j$ are the random scaled correspond pairs, while $x_i$ and $y_k$ are not correspond pairs. The trained feature extractors of AMCNet have the following properties: the extracted music feature is closely synchronized to the target motion, and the extracted motion feature is scale-invariant.
    
\begin{small} 
    \begin{equation}
        \begin{aligned}
        L_{AMC} = &\frac{1}{N}\sum_{i,j=0}^N{(\text{AMCNet}(x_i,y_j) - 1)^2}+\\ & \frac{1}{N}\sum_{i,k=0}^N{(\text{AMCNet}(x_i,y_k)-0)^2}
        \label{lossAMC}
        \end{aligned}
    \end{equation}
\end{small}

\subsection{Adversarial-Perceptual Learning}
    Due to it is difficult to learn the one-to-many mapping by using standard MSE loss, we relax the constraint of MSE by combining adversarial loss and perceptual loss. The adversarial loss enables the model to approximate the distribution of the real conductor motion and generate a realistic motion sequence, while the perceptual loss ensures the generated motion conforms to music. As shown in Fig.\ref{fig:model_structure}, we use the learned motion encoder $E_m$ of AMCNet to calculate the perceptual loss, and transfer the music encoder $E_a$ into the motion generator $G$. At the same time, a discriminator $D_m$ with Lipschitz constraint is set up to guide the generator towards the real motion distribution. The loss function of the motion generator $G$ is shown in Eq.\ref{lossG}

\begin{small} 
\begin{equation}
    \begin{aligned}
    L_{G} = &\lambda_{mse}\frac{1}{M}\sum_{i=0}^M{(G(x_i) - y_i)^2}+\\
    &\lambda_{per}\frac{1}{M}\sum_{i=0}^M{(E_m(G(x_i))-E_m(y_i))^2}+\\
    &\lambda_{adv}\frac{1}{M}\sum_{i=0}^M{D(G(x_i))}
    \label{lossG}
    \end{aligned}
\end{equation}
\end{small}

\begin{figure}
    \centering
    \includegraphics[width=0.4\textwidth]{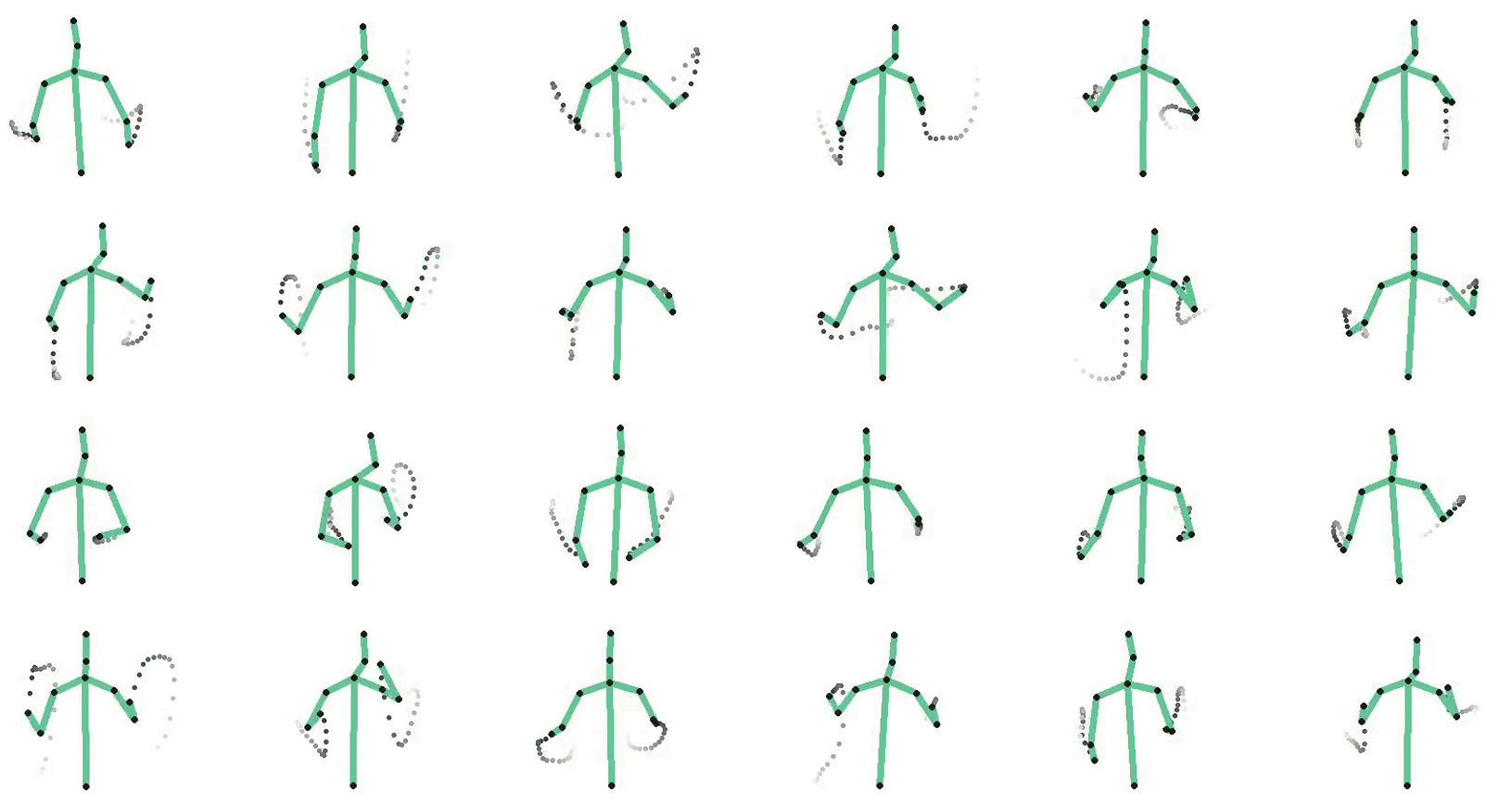}
    \caption{Skeleton generated by the \textit{VirtualConductor}.}
    \label{fig:Skeleton}
\end{figure}

\section{Demonstration}

\begin{figure}
    \centering
    \includegraphics[width=0.44\textwidth]{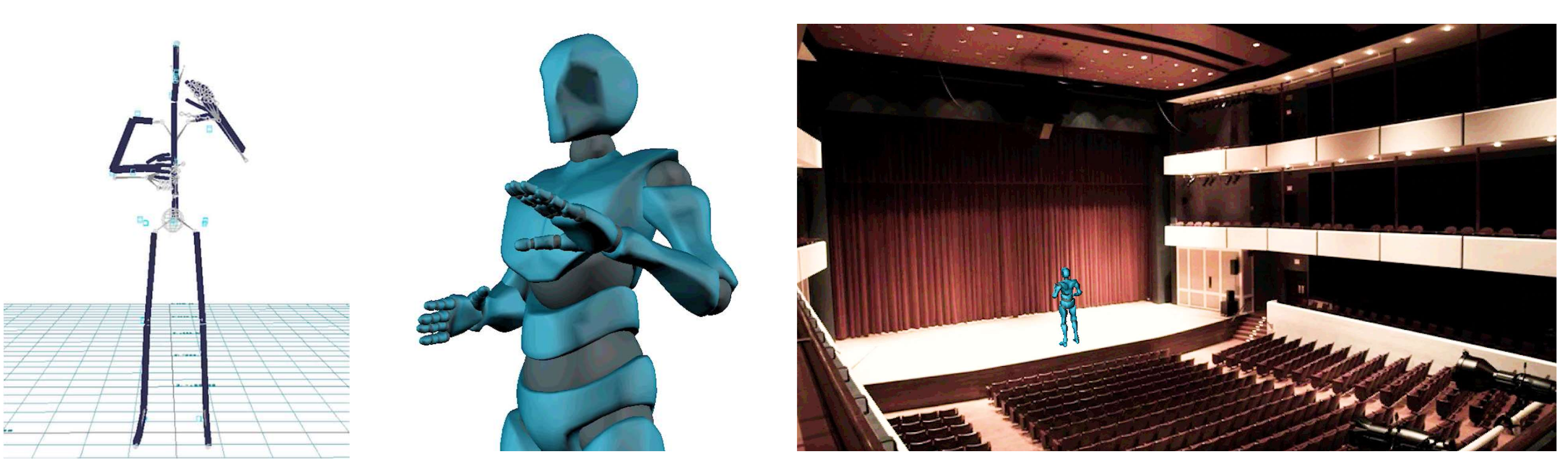}
    \caption{3D animation results generated by the \textit{VirtualConductor}.}
    \label{fig:Robot}
\end{figure}

\begin{figure}
    \centering
    \includegraphics[width=0.44\textwidth]{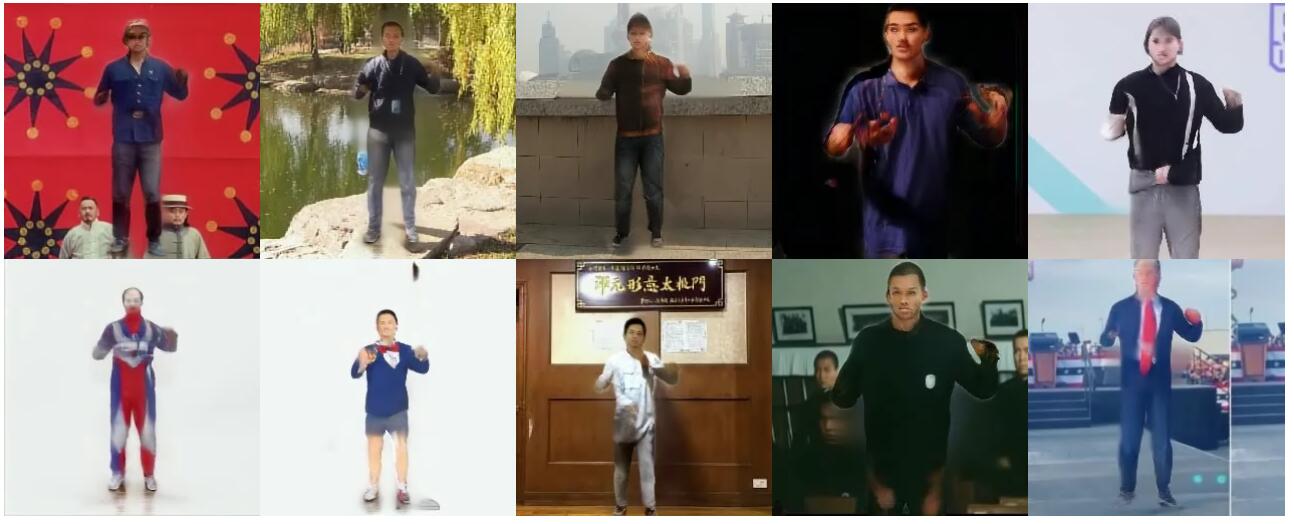}
    \caption{Demo results generated by the \textit{VirtualConductor}.}
    \label{fig:results}
\end{figure}

    The \textit{VirtualConductor} requires a music file and a single user's image as input. The system extracts the music feature and generates a motion sequence. The motion sequence is subsequently rendered to the 3D avatar, and the user's conducting video by pose transfer module \cite{LiquidWarpingGAN}. The generated skeleton, 3D animation and final video output results are respectively shown in Fig.~\ref{fig:Skeleton}, Fig.~\ref{fig:Robot}, and Fig.~\ref{fig:results}. The demo system is implemented in Pytorch with an NVIDIA 2080Ti GPU. It can generate conductor motion of complete Beethoven Symphony No.5 in 1.431 seconds and synthesize conducting video in about 13 fps.
    
\begin{figure}
    \centering
    \includegraphics[width=0.475\textwidth]{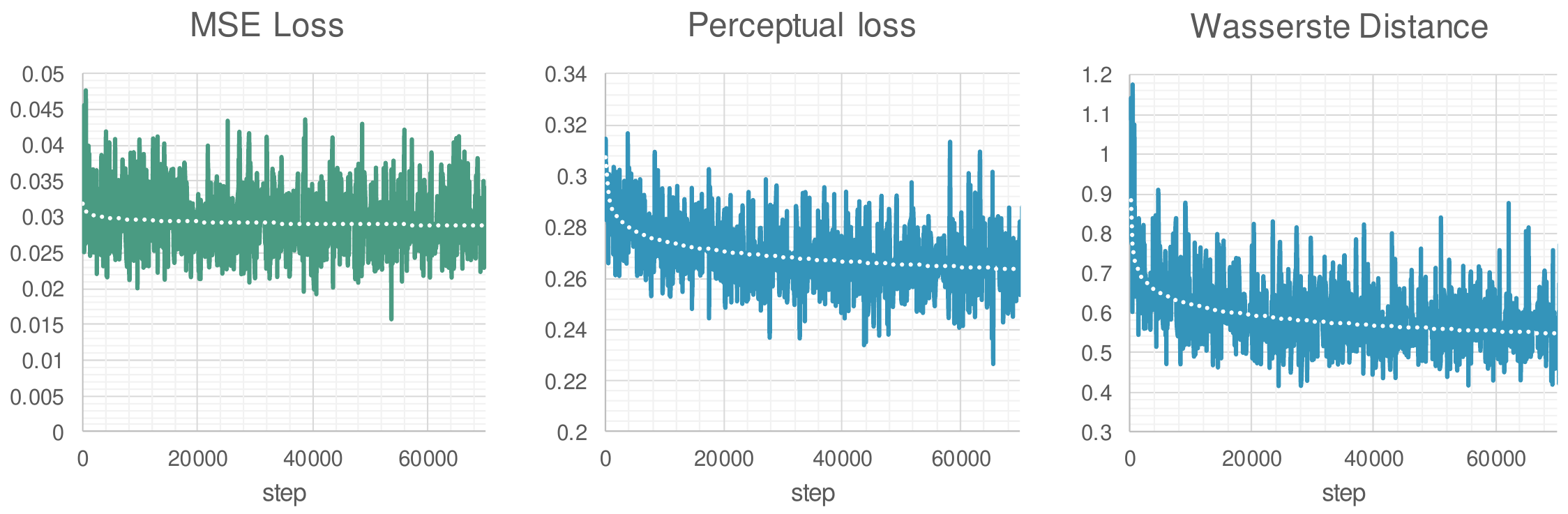}
    \caption{Loss curves and the Wasserstein distance during training.}
    \label{fig:curves}
\end{figure}

    Fig.\ref{fig:curves} shows the loss curves and the Wasserstein distance during training. Although the MSE loss has almost not been optimized, the perceptual loss and Wasserstein distance are going down. It shows that the generator is not trying to regress the ground truth motion. Instead, it learns the distribution of the real motion and its relationship with music.

\section{Conclusion}
    In this demo, we make the first attempt towards music-driven conductor motion generation. First, a large-scale conductor motion dataset is constructed. Then, we propose a 2-stage model which includes audio motion correspondence learning and adversarial perceptual learning. Experimental results show that the cross-modal relationship between music and motion is effectively learned. Finally, taking a music file and a single user's image as input, the \textit{VirtualConductor} system can generate diverse, plausible, music-synchronized conducting video and enable anyone to become a conductor.

\section{Acknowledgement}
    This work was partially funded by the Natural Science Foundation of Jiangsu Province under Grant No. BK20191298, Fundamental Research Funds for the Central Universities under Grant No. B200202175.

\bibliographystyle{IEEEbib}
\bibliography{icme2021template}

\end{document}